# Hybrid Noise Removal in Hyperspectral Imagery with a Spatial-Spectral Gradient Network

Qiang Zhang, *Student Member, IEEE*, Qiangqiang Yuan, *Member, IEEE*, Jie Li, *Member, IEEE*, Xinxin Liu, Huanfeng Shen, *Senior Member, IEEE*, Liangpei Zhang, *Fellow, IEEE*

***Abstract*—The existence of hybrid noise in hyperspectral images (HSIs) severely degrades the data quality, reduces the interpretation accuracy of HSIs, and restricts the subsequent HSIs applications. In this paper, the spatial-spectral gradient network (SSGN) is presented for mixed noise removal in HSIs. The proposed method employs a spatial-spectral gradient learning strategy, in consideration of the unique spatial structure directionality of sparse noise and spectral differences with additional complementary information for effectively extracting intrinsic and deep features of HSIs. Based on a fully cascaded multi-scale convolutional network, SSGN can simultaneously deal with the different types of noise in different HSIs or spectra by the use of the same model. The simulated and real-data experiments undertaken in this study confirmed that the proposed SSGN outperforms at mixed noise removal compared with the other state-of-the-art HSI denoising algorithms, in evaluation indices, visual assessments, and time consumption.**

## I. Introduction

DUE to the abundant spectral information, hyperspectral image (HSI) data [1] have been successfully applied in ground object classification [2], endmember extraction [3], and unmixing [4]. Nevertheless, because of the sensor instability and atmospheric interference, HSIs often suffer from multiple types of noise [5], such as Gaussian noise, stripe noise, impulse noise, dead lines, and mixed noise, as illustrated in Fig. 1. The degraded information often disturbs and limits the subsequent processing. Therefore, noise reduction in HSIs is crucial before image interpretation and the subsequent applications [6-7].

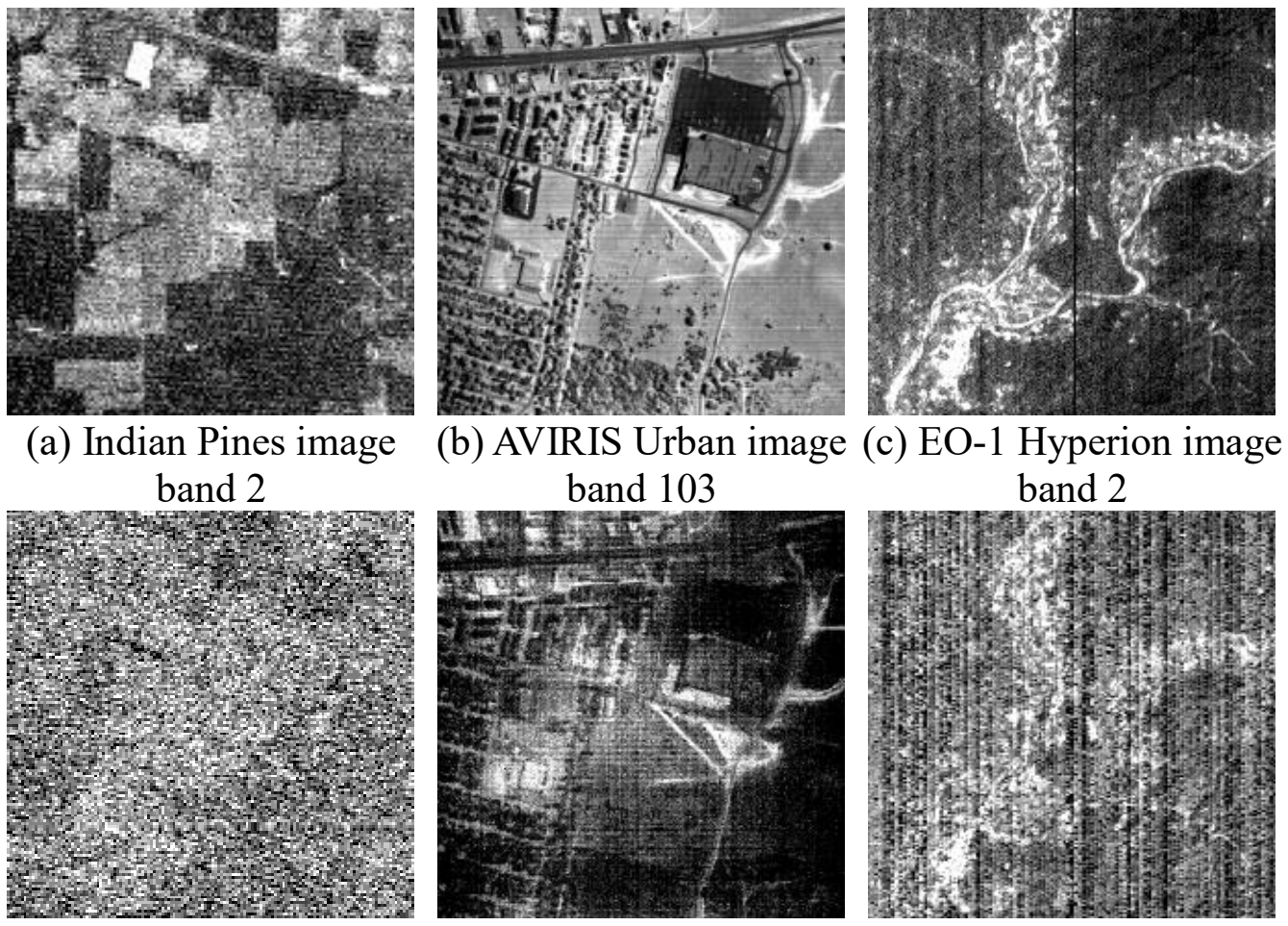

(a) Indian Pines image band 2 (b) AVIRIS Urban image band 103 (c) EO-1 Hyperion image band 2

(d) Indian Pines image band 109 (e) AVIRIS Urban image band 104 (f) EO-1 Hyperion image band 166

Fig. 1. Multiple types and levels of noise in different HSIs and different bands.

To date, by treating the HSI as 3-D cube data, many different algorithms [8-17] have been proposed for HSI denoising. Details of the main HSI denoising methods are provided in Section II. Although the existing HSI denoising methods can obtain favorable outcomes, there are still several challenges and bottlenecks that need to be solved. Firstly, the manual parameters must be adjusted suitably and carefully for different HSI data, which brings about inconvenient, non-universal and time-consuming drawbacks for different scenarios and HSI sensors. Secondly, the noise in HSIs exists in both the spatial and spectral domains, with various types and diverse levels, as displayed in Fig. 1. However, most algorithms are unable to satisfy the complex situation of mixed noise. As a result, the processed data often contain residual noise or show spectral distortion. Thirdly, most bands in HSIs are of high quality, while only some specific bands are degraded by diverse noise. Therefore, how to effectively reduce the noise in the corrupted bands while simultaneously preserving the high-quality bands is of great importance for HSI denoising. Overall, there is a need to establish a convenient, universal, efficient, and robust denoising framework that can adapt to different HSI data with

Manuscript received January 14, 2019; revised March 15, 2019; accepted March 26, 2019. This work was supported in part by the National Key Research and Development Program of China under Grant 2016YFC0200900 and 2016YFB0501403, in part by the National Natural Science Foundation of China under Grant 41701400, and in part by the Fundamental Research Funds for the Central Universities under Grant 2042017kf0180 and 531118010209.
(Corresponding authors: *Jie Li and Qiangqiang Yuan.*)

Q. Zhang is with the State Key Laboratory of Information Engineering in Surveying, Mapping and Remote Sensing, Wuhan University, Wuhan 430079, China (e-mail: whuqzhang@gmail.com).

Q. Yuan is with the School of Geodesy and Geomatics and the Collaborative Innovation Center of Geospatial Technology, Wuhan University, Wuhan 430079, China (e-mail: qqyuan@sgg.whu.edu.cn).

J. Li is with the School of Geodesy and Geomatics, Wuhan University, Wuhan 430079, China (e-mail: aaronleecool@whu.edu.cn).

X. Liu is with the College of Electrical and Information Engineering, Hunan University, Changsha 410082, China (e-mail: liuxinxin@hnu.edu.cn)

H. Shen is with the School of Resource and Environmental Science and the Collaborative Innovation Center of Geospatial Technology, Wuhan University, Wuhan 430079, China (e-mail: shenhf@whu.edu.cn).

L. Zhang is with the State Key Laboratory of Information Engineering in Surveying, Mapping and Remote Sensing and the Collaborative Innovation Center of Geospatial Technology, Wuhan University, Wuhan 430079, China (e-mail: zlp62@whu.edu.cn).

different and mixed noise types.

Recently, on account of the powerful feature extraction and nonlinear expression ability brought by deep learning like deep convolutional neural networks (DCNNs) [18-19], many low-level vision problems [20-23] in remote sensing data, such as SAR image despeckling [20], pansharpening [21-22], and missing data reconstruction [23], have been provided with a learning framework which can achieve a state-of-the-art performance. To overcome the drawbacks mentioned above for HSIs denoising and take full advantage of DCNNs, we propose a spatial-spectral gradient network (SSGN) for hybrid noise reduction in HSIs considering the noise type of Gaussian noise, stripe noise, impulse noise, dead line and their mixture. The main innovations can be generalized as below.

1) A spatial-spectral convolutional network is proposed for HSI denoising. To take advantage of the abundant spectral information in HSIs and the distinct spatial information for each band, SSGN simultaneously employs the spatial data and the adjacent spectral data in fully cascaded multi-scale convolutional neural network blocks.

2) The spatial gradient and spectral gradient are jointly incorporated in the proposed model. The spatial gradient is utilized to extract the unique structure directionality of sparse noise in the horizontal and vertical directions, and the spectral gradient is used to obtain spectral additional complementary information for the noise removal. The spectral gradient is also integrated into the spatial-spectral loss function to reduce the spectral distortion in the whole framework.

3) The experimental results confirm that the proposed method can effectively deal with Gaussian noise, stripe noise, and mixed noise in different HSIs or spectra through single model. Compared with the other state-of-the-art HSI denoising algorithms, SSGN outperforms in evaluation indices, visual assessments, and time consumption, under different mixed noise scenarios.

The rest of this paper is organized as follows. Section II describes the HSI degradation procedure caused by hybrid noise, and then introduces the existing HSI denoising methods. In Section III, the proposed model is described. The simulated and real-data experimental results are presented in Section IV. Finally, our conclusions are given in Section V.

## II. Related Work

Given an HSI is a three-dimensional tensor $\mathbf{Y} \in \mathbb{R}^{M \times N \times B}$, where $M$ and $N$ represent the spatial dimension and $B$ denotes the spectral dimension, the HSI noise degradation model can be described as [24]:

$$\mathbf{Y} = \mathbf{X} + \mathbf{D} + \mathbf{S} \tag{1}$$

where $\mathbf{X} \in \mathbb{R}^{M \times N \times B}$ represents clean HSI data, $\mathbf{D} \in \mathbb{R}^{M \times N \times B}$ denotes dense noise such as Gaussian noise, and $\mathbf{S} \in \mathbb{R}^{M \times N \times B}$ denotes sparse noise such as sparse-distributed stripe noise and dead lines. To obtain noise-free data $\mathbf{X}$ from Eq. (1) with only $\mathbf{Y}$ known, many scholars have developed different models for the HSI denoising problem. The existing HSI denoising methods can be roughly classified into two types [26]: 1) *filter-based methods*; and 2) *model optimization-based methods*. The specific peculiarities, advantages, and disadvantages of these two types of methods are described as follows.

***1) Filter-Based Methods:*** The filter-based methods aim to separate clean signals from the noisy signals through filtering operations, such as Fourier transform, wavelet transform, or a non-local means (NLM) filter. For example, Othman *et al.* [8] proposed a combined spatial-spectral derivative-domain wavelet shrinkage noise removal method. This method depends on the spectral derivative domain, and benefits from the dissimilarity of the signal nature in the spatial and spectral dimensions where the noise level is elevated. In addition, based on the NLM strategy, Chen *et al.* [9] presented an extension of the block-matching and 3D filtering (BM3D) [27] algorithm from two-dimensional data to a three-dimensional data cube, employing principal component analysis (PCA) and 3-D transform for the noise reduction. The major drawback of these filtering methods lies in the usage of the handcrafted and fixed wavelet, which are sensitive to the selection of the transform function and cannot consider the differences in the geometrical characteristics of HSIs such as mixed noise.

***2) Model Optimization-Based Methods:*** The model optimization based methods, such as total variation [10], sparse representation [11]–[12], and low-rank matrix and tensor models [13]–[17], take the reasonable assumption or priors of the HSI data into account. This type of method can map the noisy HSI to the clear one in an attempt to preserve the spatial and spectral characteristics. For example, considering the noise intensity difference in different bands, Yuan *et al.* [10] proposed a spatial-spectral adaptive total variation denoising algorithm. Furthermore, by applying the sparsity prior of the HSI data, Zhao *et al.* [11] investigated sparse coding to describe the global redundancy and correlation (RAC) and the local RAC in the spectral domain, and then employed a low-rank constraint to capture the global RAC in the spectral domain. Furthermore, Li *et al.* [12] exploited the intra-band structure and the inter-band correlation in the process of joint sparse representation and joint dictionary learning.

For an HSI, both the high spectral correlation between adjacent bands and the high spatial similarity within one band can reveal the low-rank property [13-14] or tensor [15-17] structure of the HSI. Hence, by lexicographically transforming a 3-D cube into a 2-D matrix representation along the spectral dimension, Zhang *et al.* [13] and He *et al.* [14] proposed a low-rank matrix restoration model for mixed noise removal in HSIs. Recently, to adequately utilize the spectral-spatial structural property for the 3-D tensor HSIs, low-rank tensor-based HSI denoising methods [15]–[17] have been proposed and have achieved state-of-the-art performances, at the cost of higher computational time consumption.

In summary, although the existing HSI denoising methods can obtain favorable results, ***the inadaptability for hybrid noise removal in different HSIs and the low efficiency issue still restrict the application of HSI denoising***. Therefore, to overcome the deficiency of the above-mentioned methods to some degree, and take full advantage of a DCNN in remote sensing data [28-30], we present the SSGN model for efficient hybrid noise reduction in HSIs.

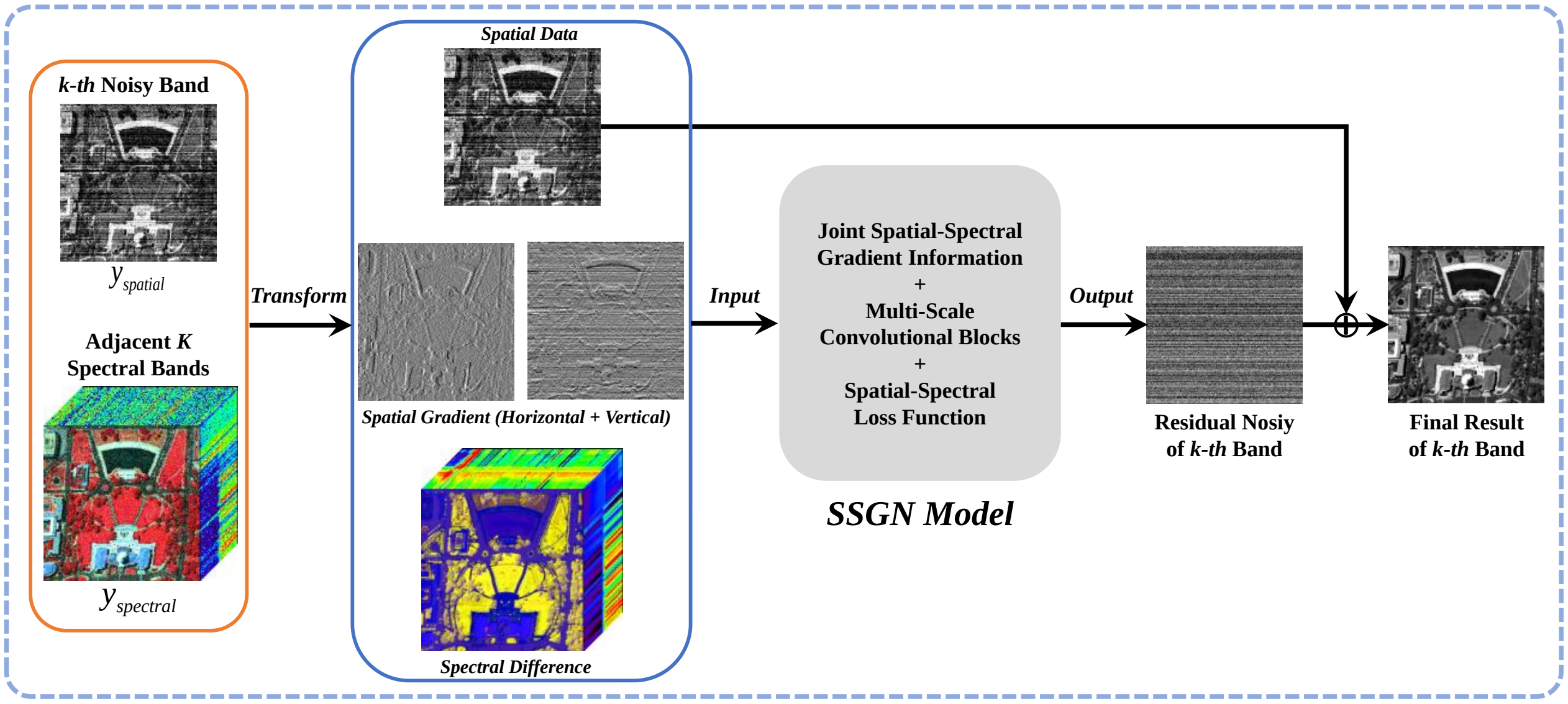

Fig. 2. Flowchart of the HSI denoising procedure with the proposed method.

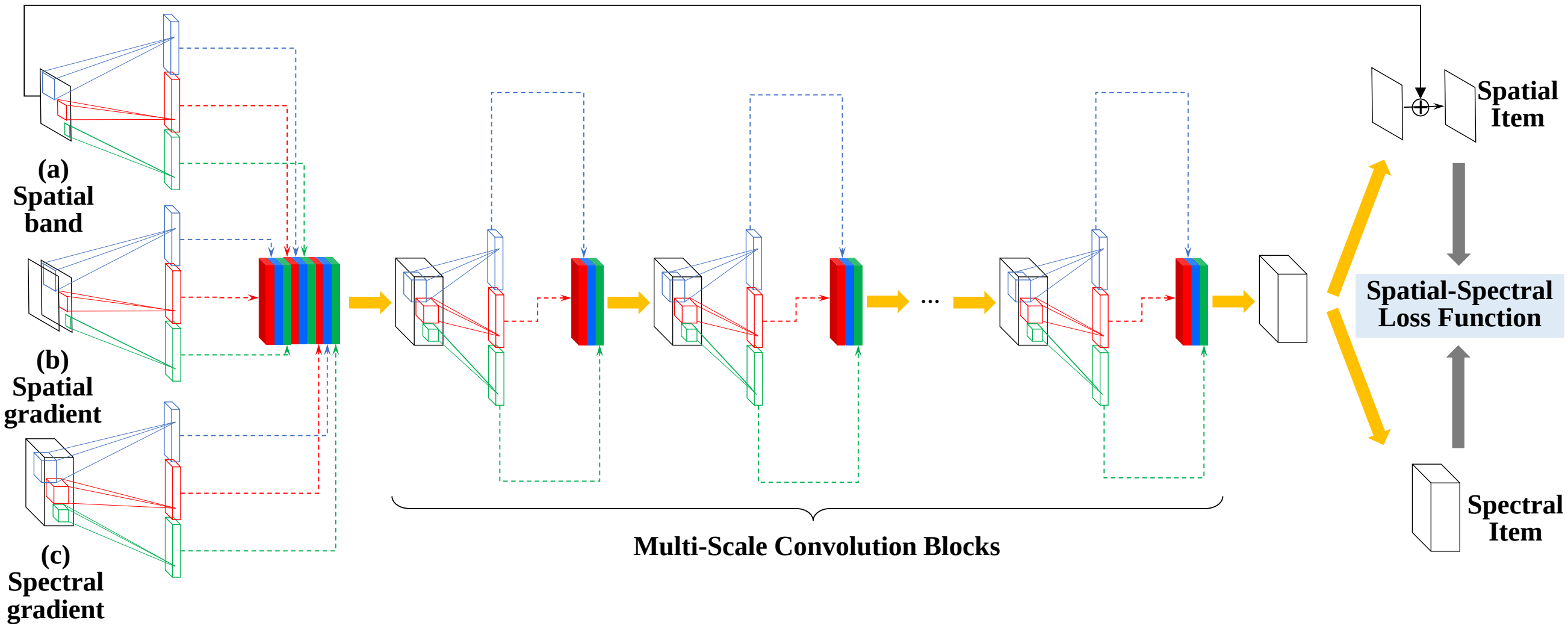

Fig. 3. The proposed SSGN model.

## III. Methodology

### A. Holistic Framework Description

To remove the diverse types of noise in HSIs, the proposed SSGN method takes the noise structure characteristic, the spatial peculiarity for each band, and the spectral redundancy into account. Through learning in an end-to-end fashion between a noisy HSI patch and a clean HSI patch, we present the SSGN model for HSI hybrid noise reduction. The SSGN model simultaneously employs the simulated $k$-th noisy band, its horizontal/vertical spatial gradient, and its adjacent spectral gradient as the input data, which outputs the residual noise of the $k$-th noisy band. Then, by traversing all the bands of the HSI in this way, we can finally obtain the denoising results for all the bands. The flowchart of the HSI denoising procedure with the SSGN model is depicted in Fig. 2. Specific details of the proposed SSGN model are provided in Section III-B.

### B. The Proposed SSGN Model for HSI Denoising

Fig. 3 illustrates the architecture of the proposed SSGN model. The input spatial band represents the current noisy band in the top-left corner. The spatial gradient in the middle-left corner denotes the vertical and horizontal gradient of the input spatial band. Correspondingly, the input spectral gradient represents the adjacent spectral difference cube with the current spatial noisy band in the bottom-left corner. Based on an end-to-end framework with fully cascaded multi-scale convolutional blocks, the proposed SSGN employs a spatial-spectral loss function to optimize the model's trainable parameters. The proposed method then traverses all the spatial bands and their adjacent spectral bands in the HSI, which simultaneously utilizes the spatial-spectral gradient information for HSI denoising.

#### 1) Joint Spatial and Spectral Gradient Information

To some extent, the gradient information of the spatial band can effectively highlight the sparse noise, especially sparse-distributed stripe noise, because of its unique structure directionality, as shown in Fig. 4. Furthermore, as HSI data contain abundant spectral information with hundreds of bands, the noise levels and types in each band are usually different. These differences in HSIs provide additional complementary information, which can be beneficial to remove mixed noise for HSIs. From the above, we argue that mixed noise in HSIs, including dense noise and spare noise, can be removed from both the spatial and spectral domains by employing joint spatial and spectral gradient information, as follows:

$$\mathbf{G}_x(m,n,k) = \mathbf{Y}(m+1,n,k) - \mathbf{Y}(m,n,k) \tag{2}$$

$$\mathbf{G}_y(m,n,k) = \mathbf{Y}(m,n+1,k) - \mathbf{Y}(m,n,k) \tag{3}$$

$$\mathbf{G}_z(m,n,k) = \mathbf{Y}(m,n,k+1) - \mathbf{Y}(m,n,k) \tag{4}$$

where $\mathbf{G}_x$, $\mathbf{G}_y$, and $\mathbf{G}_z$ stand for the horizontal spatial gradient, the vertical spatial gradient, and the spectral gradient of the current band $\mathbf{Y}_k$, respectively. For the $k$-th band in the HSI, $\mathbf{G}_z$ represents its adjacent bands with the number $K$. Instead of directly generating the final denoising results, a residual learning strategy is utilized to estimate the noise elements, which can also ensure the stability and efficiency of the training procedure in the proposed SSGN model [31-33]. The final reconstruction output of $k$-th band is denoted as:

$$\hat{\mathbf{X}}_k = \mathbf{F}_l \circ \ldots \circ \mathbf{F}_2 \circ \mathbf{F}_1 \left\{ \mathbf{Y}_k, \mathbf{G}_x, \mathbf{G}_y, \mathbf{G}_z \right\} + \mathbf{Y}_k \tag{5}$$

where $\mathbf{F}_l$ represents the $l$-th multi-scale convolutional block operation of the proposed model, $\circ$ stands for the feature map transformation from $\mathbf{F}_{l-1}$ to $\mathbf{F}_l$, and $\hat{\mathbf{X}}_k$ represents the estimated denoising result of the $k$-th band. The multi-scale convolutional blocks are introduced in detail below.

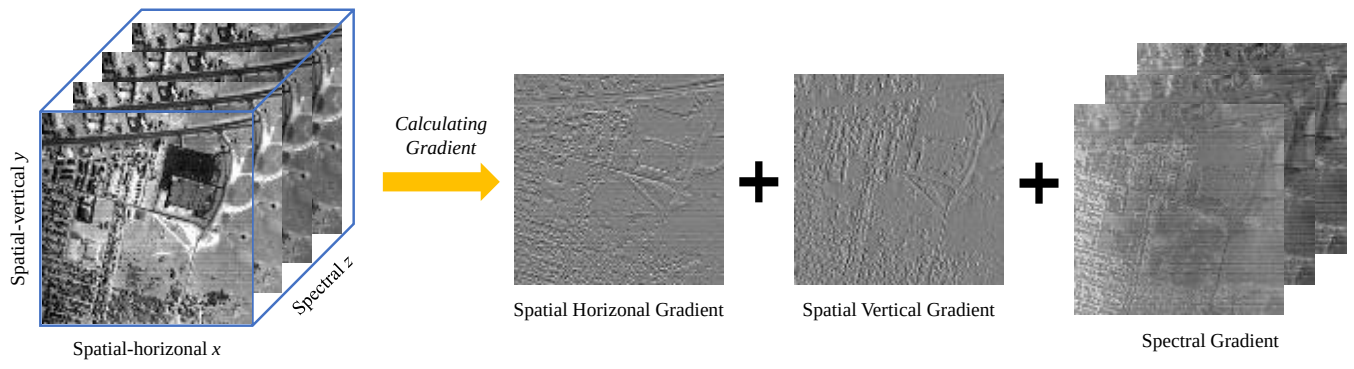

Fig. 4. The spatial and spectral gradient results in the HSI.

#### 2) Multi-Scale Convolutional Blocks

In HSI data, as shown in Fig. 5(a), the feature expression may count on contextual information in different scales, since ground objects usually have multiplicative sizes in different non-local regions. Furthermore, multi-scale convolutional filters can simultaneously obtain diverse receptive field sizes, especially for the scenario with stripe noise and dead lines, as displayed in Fig. 5(b). From this perspective, to effectively eliminate sparsely distributed noise such as stripe noise and dead lines in HSIs, the proposed SSGN model introduces multi-scale convolutional blocks to extract multi-scale features for the multi-context information. In addition, to capture both the multi-scale spatial feature and spectral feature in HSIs, the proposed method employs different convolutional kernel sizes, as described in Fig. 5. The multi-scale convolutional blocks contain three convolution operations of $3\times3$ (green), $5\times5$ (red), and $7\times7$ (blue) kernel sizes for the spatial data and spectral data, respectively. All these three convolutions for the current spatial band, spatial gradient, and corresponding spectral gradient produce feature maps of 30 channels, as revealed in Fig. 3(a)–(c), respectively.

In addition, the proposed SSGN also employs fully cascaded multi-scale convolutional blocks for extracting more feature maps with different receptive field sizes. As shown in Fig. 5(c), as the depth of the layers increases, the results of the different blocks gradually approximate to the final residual mixed noise, including dense noise such as Gaussian noise, and sparse noise such as sparse-distributed stripe noise and dead lines.

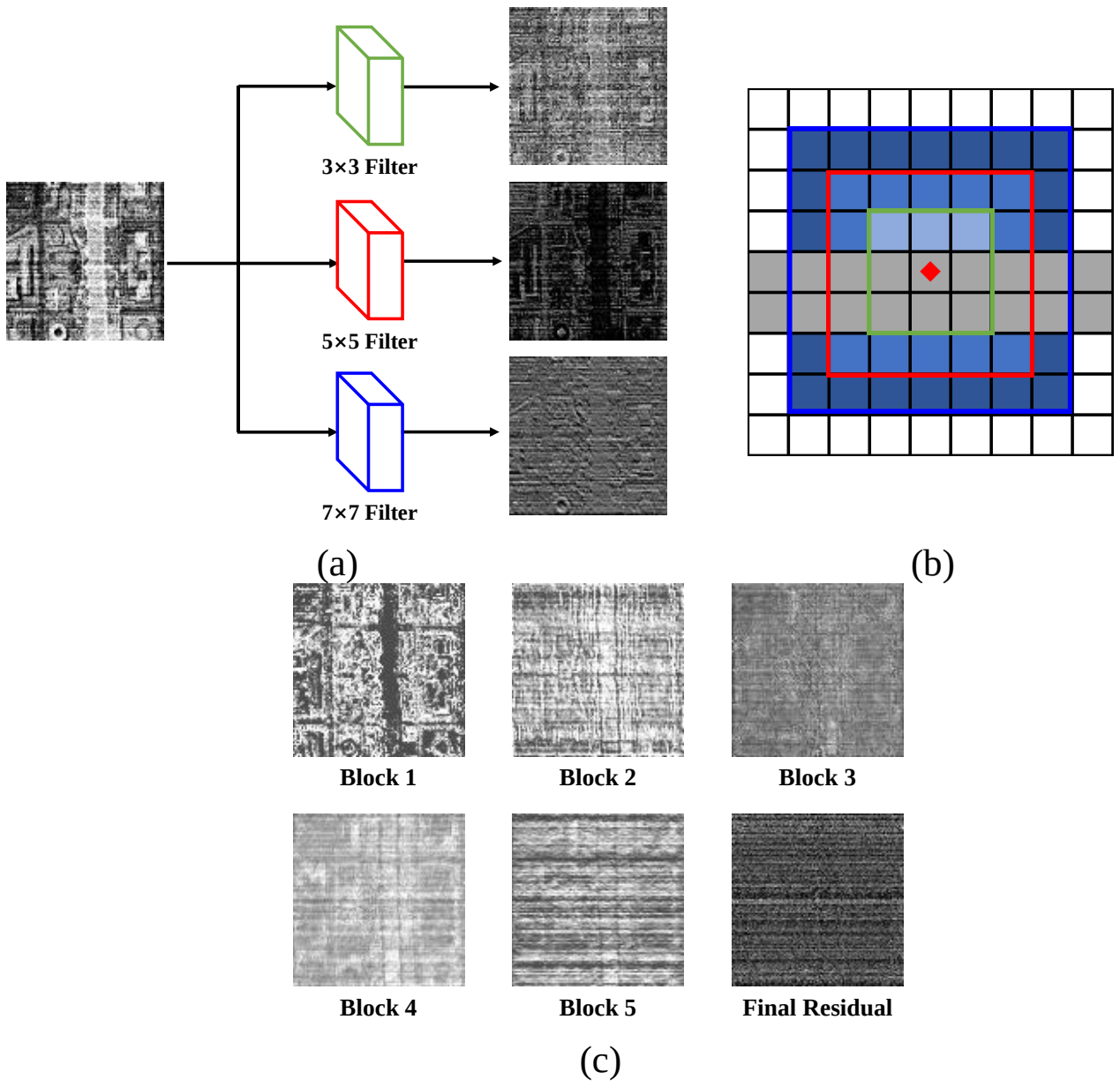

Fig. 5. (a) Multi-scale convolutional block in Fig. 2. (b) Receptive field with sparse noise, such as sparse-distributed stripes or dead lines. (c) Feature maps through the different cascaded blocks.

#### 3) Spatial-Spectral Loss Function

Loss functions form an indispensable module in the supervised learning procedure. To optimize the model parameters, the spatial-spectral loss function in the proposed SSGN method is utilized in the training process with a back-propagation algorithm. The traditional image restoration tasks, such as super-resolution and denoising, usually utilize a Euclidean loss function, which only considers the spatial information restoration, and not the spectral information. Meanwhile, in HSI restoration, spectral preservation must be considered, which is crucial for the subsequent applications such as unmixing. Therefore, to simultaneously maintain the spatial structure information and restrain the spectral distortion, the proposed method develops a spatial-spectral loss function during the training procedure, as follows:

$$\xi(\Theta) = (1-\alpha)\cdot\xi_{spatial} + \alpha\cdot\xi_{spectral} \tag{6}$$

where $\xi_{spatial}$ (spatial term) and $\xi_{spectral}$ (spectral term) are respectively defined as:

$$\xi_{spatial} = \frac{1}{2T}\sum_{i=1}^{T}\left\|\mathbf{Res}_k^i - (\mathbf{Y}_k^i - \mathbf{X}_k^i)\right\|_2^2 \quad (7)$$

$$\xi_{spectral} = \frac{1}{2T}\sum_{i=1}^{T}\sum_{z=k-\frac{K}{2}, z\neq k}^{k+\frac{K}{2}}\left\|\Phi_z^i - \mathbf{G}_z^i\right\|_2^2 \quad (8)$$

where $\alpha$ is the penalty parameter for the trade-off between the spatial and spectral items, $T$ stands for the number of training patches, $\mathbf{Res}_k$ is the estimated residual spatial output of the $k$-th band, and $\Phi_z$ represents the estimated spectral gradient of bands $[k-\frac{K}{2}:k-1, k+1:k+\frac{K}{2}]$. To sufficiently elaborate the effectiveness of the spatial-spectral loss function, a discussion is provided in Section IV-C.

## IV. Experiments And Discussions

To validate the effectiveness of the proposed SSGN for hybrid noise reduction in HSIs, both simulated and real noisy HSI data were employed. The proposed method was compared with the five existing state-of-the-art methods of hybrid spatial-spectral noise reduction (HSSNR) [8], block-matching and 4D filtering (BM4D) [9], low-rank matrix recovery (LRMR) [13], low-rank total variation (LRTV) [14], and nonconvex low-rank matrix approximation (NonLRMA) [34]. Before the denoising procedure, the gray values of each HSI band were independently all normalized to [0-1]. The mean peak signal-to-noise ratio (MPSNR) [35], the mean structural similarity index (MSSIM) [36], and the mean spectral angle (MSA) [37] were employed as assessment indices in the simulated experiments. Generally speaking, in simulated experiments, better HSI denoising effects can be reflected by higher MPSNR and MSSIM values and lower MSA values. For the real-data experiments, the three HSIs were tested and the reconstruction mean digital number (DN) results of all the rows or columns are displayed. The code of the proposed method can be downloaded from https://github.com/WHUQZhang.

**1) Parameter Settings:** The adjacent spectral band number was set as $K = 24$ and the trade-off parameter was set as $\alpha = 0.001$ in the SSGN model for both the simulated and real-data experiments. An impact analysis for this parameter is provided in the discussion part. The proposed SSGN model was trained with the Adam optimization algorithm [38] as the gradient descent optimization method, with momentum parameters of 0.9, 0.999, and $10^{-8}$, respectively. In addition, the learning rate was initialized to 0.001 for the whole network. After every 10 epochs, the learning rate was decreased through being multiplied by a descent factor of 0.5.

**2) Network Training:** For the training of the proposed SSGN model, the University of Pavia image obtained by the airborne Reflective Optics System Imaging Spectrometer (ROSIS) sensor, with the size of $610\times340\times96$, and the Washington DC Mall image obtained by the Hyperspectral Digital Imagery Collection Experiment (HYDICE) airborne sensor, with the size of $1080\times303\times191$, were both employed as training data after removing the noisy and water absorption bands. These training labels were then cropped in each patch size as $25\times25$, with the sampling stride equaling to 25. The simulated mixed noise patch data were generated through imposing additive white Gaussian noise (AWGN), stripe noise, and dead lines on the different spectra. The noise intensity and degree of distribution were also varied in the different bands. Due to the fact that an increasing number of HSI training samples can effectively fit the HSI denoising model, multi-angle image rotation (angles of 0, 90, 180, and 270 degrees in our training data sets) and multi-scale resizing (scales of 0.8, 1, 1.2, and 1.4 in our training data sets) were both utilized during the training procedure. The training process of SSGN took 200 epochs (an epoch is equal to about 1,200 iterations). We employed the Caffe framework [39] to train the proposed model on a PC with Windows 7 environment, 16 GB RAM, an Intel Xeon E5-2609 v3 CPU, and an NVIDIA Titan X GPU. The training process for the proposed model cost roughly 13 h 45 mins.

**3) Test Data Sets:** Four data sets were employed in the simulated and real-data experiments, as follows.

a) The first data set was the Washington DC Mall image mentioned above in Section IV-B, which was cropped to $200\times200\times191$ for the simulated-data experiments, after removing the water absorption bands.

b) The second data set was the HYDICE Urban HSI with the size of $307\times307\times188$, which was employed for the real-data experiments after removing the severely degraded bands.

c) The third data set was the Airborne Visible InfraRed Imaging Spectrometer (AVIRIS) Indian Pines HSI with the size of $145\times145$, which was employed for the real-data experiments. A total of 206 bands were utilized in the experiments after removing bands 150–163, which are severely disturbed by the atmosphere and water.

d) The fourth data set was the EO-1 Hyperion data set with the size of $400\times200\times166$, which was employed for the real-data experiments after removing the water absorption bands.

### A. Simulated-Data Experiments

In the simulated HSI hybrid noise reduction process, the additional noise was simulated as the following five cases.

***Case 1 (Gaussian noise):*** All bands in the HSIs were corrupted by Gaussian noise. For the different spectra, the noise intensity was different and conformed to a random probability distribution [40].

***Case 2 (stripe noise):*** A part of the bands in the HSIs were corrupted by stripe noise. In our experiments, 10 bands of the original data were imposed with simulated stripe noise [41-42]. The simulated stripe strategy is that we randomly select some rows in original bands. Half of these rows are integrally aggrandized with the strength of mean, and the remaining half are reduced with the strength of mean.

***Case 3 (Gaussian noise + stripe noise):*** All bands in the HSIs were corrupted by Gaussian noise and some of the bands were corrupted by stripe noise. The Gaussian noise intensity and stripe noise were identical to ***Case 1*** and ***Case 2***, respectively.

TABLE I

QUANTITATIVE EVALUATION OF THE HYBRID NOISE REDUCTION RESULTS IN THE SIMULATED EXPERIMENTS

| | Noisy HSI | HSSNR | BM4D | LRMR | LRTV | NonLRMA | SSGN |
|---|---|---|---|---|---|---|---|
| ***Case 1: Gaussian noise*** | | | | | | | |
| MPSNR | 23.27 | 27.25 | 28.62 | 33.21 | 33.96 | 32.15 | **34.37** |
| MSSIM | 0.769 | 0.923 | 0.941 | 0.981 | 0.980 | 0.981 | **0.982** |
| MSA | 19.47 | 9.083 | 5.116 | 4.628 | 5.507 | 5.312 | **4.241** |
| Time/s | - | 304.4 | 461.8 | 449.6 | 175.1 | 548.4 | **7.3** |
| ***Case 2: Stripe noise*** | | | | | | | |
| MPSNR | 32.24 | 34.34 | 33.49 | 37.68 | 36.17 | 37.89 | **39.15** |
| MSSIM | 0.945 | 0.967 | 0.959 | 0.974 | 0.973 | 0.976 | **0.992** |
| MSA | 7.132 | 6.324 | 6.814 | 5.175 | 5.351 | 4.862 | **2.904** |
| Time/s | - | 319.7 | 445.8 | 527.4 | 169.9 | 610.7 | **7.5** |
| ***Case 3: Gaussian noise + stripe noise*** | | | | | | | |
| MPSNR | 21.66 | 26.08 | 28.24 | 31.27 | 31.74 | 31.28 | **33.18** |
| MSSIM | 0.745 | 0.903 | 0.935 | 0.958 | 0.955 | 0.976 | **0.979** |
| MSA | 20.57 | 10.45 | 8.285 | 7.202 | 7.992 | 5.598 | **4.485** |
| Time/s | - | 301.1 | 423.3 | 521.7 | 180.1 | 551.6 | **7.4** |
| ***Case 4: Gaussian noise + dead lines*** | | | | | | | |
| MPSNR | 21.91 | 25.94 | 27.49 | 31.31 | 32.06 | 30.65 | **33.86** |
| MSSIM | 0.752 | 0.908 | 0.931 | 0.967 | 0.965 | 0.971 | **0.979** |
| MSA | 20.25 | 10.063 | 5.712 | 5.732 | 6.583 | 6.041 | **4.595** |
| Time/s | - | 304.9 | 477.7 | 478.8 | 182.6 | 555.2 | **7.3** |
| ***Case 5.1: Gaussian noise + stripe noise + dead lines (SNR = 8dB)*** | | | | | | | |
| MPSNR | 21.36 | 25.63 | 27.28 | 30.79 | 31.34 | 30.29 | **32.69** |
| MSSIM | 0.735 | 0.898 | 0.927 | 0.950 | 0.948 | 0.959 | **0.975** |
| MSA | 22.34 | 10.81 | 5.857 | 7.998 | 8.209 | 7.083 | **4.681** |
| Time/s | - | 315.6 | 452.9 | 532.1 | 167.2 | 545.4 | **7.4** |
| ***Case 5.2: Gaussian noise + stripe noise + dead lines (SNR = 18dB)*** | | | | | | | |
| MPSNR | 24.15 | 28.03 | 31.48 | 33.57 | 33.63 | 34.25 | **35.14** |
| MSSIM | 0.782 | 0.937 | 0.965 | 0.974 | 0.979 | 0.982 | **0.985** |
| MSA | 17.63 | 8.145 | 4.873 | 4.465 | 5.156 | 4.367 | **4.186** |
| Time/s | - | 302.6 | 466.7 | 515.3 | 178.4 | 558.3 | **7.4** |
| ***Case 5.3: Gaussian noise + stripe noise + dead lines (SNR = 28dB)*** | | | | | | | |
| MPSNR | 31.86 | 33.68 | 34.37 | 38.48 | 37.96 | 40.36 | **40.65** |
| MSSIM | 0.937 | 0.956 | 0.964 | 0.981 | 0.978 | 0.989 | **0.991** |
| MSA | 7.842 | 6.543 | 6.468 | 5.347 | 5.357 | 3.975 | **3.458** |
| Time/s | - | 308.5 | 455.3 | 507.4 | 179.5 | 573.7 | **7.5** |
| ***Case 5.4: Gaussian noise + stripe noise + dead lines (SNR = 38dB)*** | | | | | | | |
| MPSNR | 39.16 | 40.12 | 41.07 | 43.56 | 43.26 | **45.98** | 45.36 |
| MSSIM | 0.974 | 0.979 | 0.982 | 0.991 | 0.990 | **0.994** | 0.993 |
| MSA | 4.635 | 4.561 | 4.246 | 3.378 | 3.424 | 3.175 | **2.894** |
| Time/s | - | 313.8 | 464.3 | 497.5 | 169.7 | 546.4 | **7.3** |

***Case 4 (Gaussian noise + dead lines):*** All the bands in the HSIs were corrupted by Gaussian noise and some of the bands were corrupted by dead lines. In our experiments, 20 bands of the original data were imposed with dead lines. The Gaussian noise intensity was identical to ***Case 1***.

***Case 5 (Gaussian noise + stripe noise + dead lines):*** All the bands in the HSIs were corrupted by Gaussian noise and some of the bands were corrupted by dead lines and stripe noise. Depending on the signal-to-noise ratio (SNR), This case is divided into four levels (***Case 5.1-5.4***). In ***Case 5.1*** *(SNR=8dB)*, the Gaussian noise intensity, stripes, and dead lines were identical to ***Case 1***, ***Case 2***, and ***Case 4***, respectively. Besides, the SNR values of ***Case 5.2***, ***5.3***, and ***5.4*** gradually rise with 18dB, 28dB, and 38dB simulated scenarios, respectively.

To acquire an integrated comparison for the other methods and the proposed SSGN, quantitative evaluation indices (MPSNR, MSSIM, and MSA) [43-44], a visual comparison, curves of the spectra, and the spectral difference results were used to analyze the results of the different methods. The average evaluation indices of the five cases with mixed noise are listed in Table I. To give detailed contrasting results, ***Case 3-5.1*** are chosen to demonstrate the visual results, corresponding to Fig. 6, Fig. 7, and Fig. 8, respectively. Due to the large number of bands in an HSI, only a few bands are selected to give the visual results in each case with pseudo-color or gray color. Fig. 6 shows the denoising results of the different algorithms in simulated ***Case 3*** with the pseudo-color view of bands 17, 27, and 57 (see enlarged details in the left corner of Fig. 6); Fig. 7 gives the denoising results of the different algorithms in simulated ***Case 4*** (see enlarged details in the right corner of Fig. 7); Fig. 8 shows the denoising results of the different algorithms in simulated ***Case 5.1***. The values of the peak signal-to-noise ratio (PSNR) and structural similarity index (SSIM) within the different bands of the restored HSI in ***Case 3-5.1*** are depicted to assess the per-band denoising result in Fig. 9.

In Table I, the best performance for each quality index is marked in bold. Compared with the other algorithms, the proposed SSGN model achieves the highest MPSNR and MSSIM values and the lowest MSA values in all the simulated ***Case 1-5.3***, in addition to showing a preferable visual quality in Figs. 6–8. Although the HSSNR algorithm has an effective noise reduction ability under weak noise levels, as shown in Table I for ***Case 3***, it cannot deal well with degraded bands with strong Gaussian noise, and the results still contain obvious residual noise, especially in Fig. 7. Furthermore, the stripes and dead lines are also not removed in ***Case 3-5.1*** through HSSNR. From Table I, BM4D shows a favorable noise reduction ability under the non-uniform noise intensities for different bands. However, it also produces over-smoothing in the results in Figs. 6–8, since the different non-local similar cubes in the HSI may result in the removal of small texture features. By exploring the low-rank property in spatial or spectral domain of the HSI, LRMR, LRTV, and NonLRMA provide favorable denoising results in Figs. 6-8 and ***Case 5.4*** with high level SNR value, respectively. However, there are still stripes, spectral distortion, and dead lines in the magnified areas, especially for the mixed noise, as in Fig. 8, due to the complexity of the mixed noise model in HSIs.

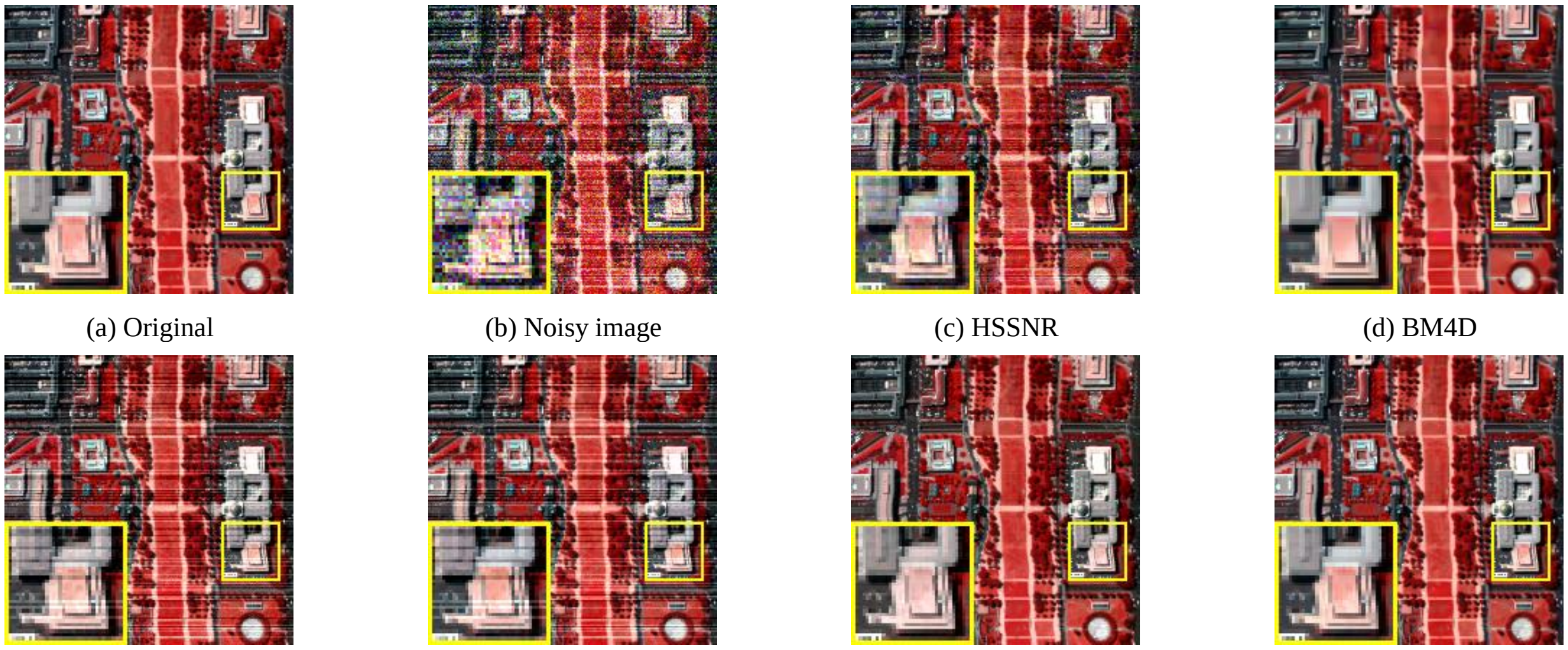

(a) Original (b) Noisy image (c) HSSNR (d) BM4D
(e) LRMR (f) LRTV (g) NonLRMA (h) SSGN

Fig. 6. *Case 3*. (a) Pseudo-color original image with bands (57, 27, 17). (b) Noisy image. (c) HSSNR. (d) BM4D. (e) LRMR. (f) LRTV. (g) NonLRMA. (h) SSGN.

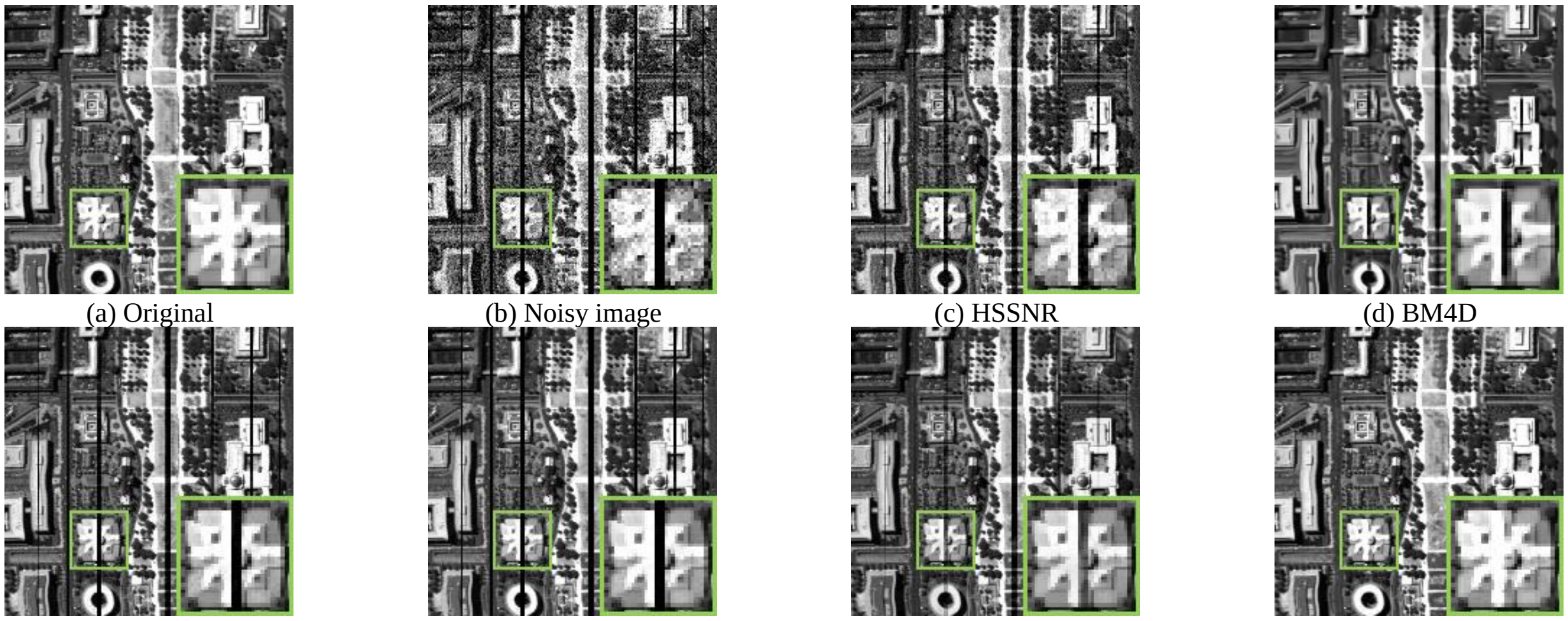

(a) Original (b) Noisy image (c) HSSNR (d) BM4D
(e) LRMR (f) LRTV (g) NonLRMA (h) SSGN

Fig. 7. *Case 4*. (a) Original image with band 164. (b) Noisy image. (c) HSSNR. (d) BM4D. (e) LRMR. (f) LRTV. (g) NonLRMA. (h) SSGN.

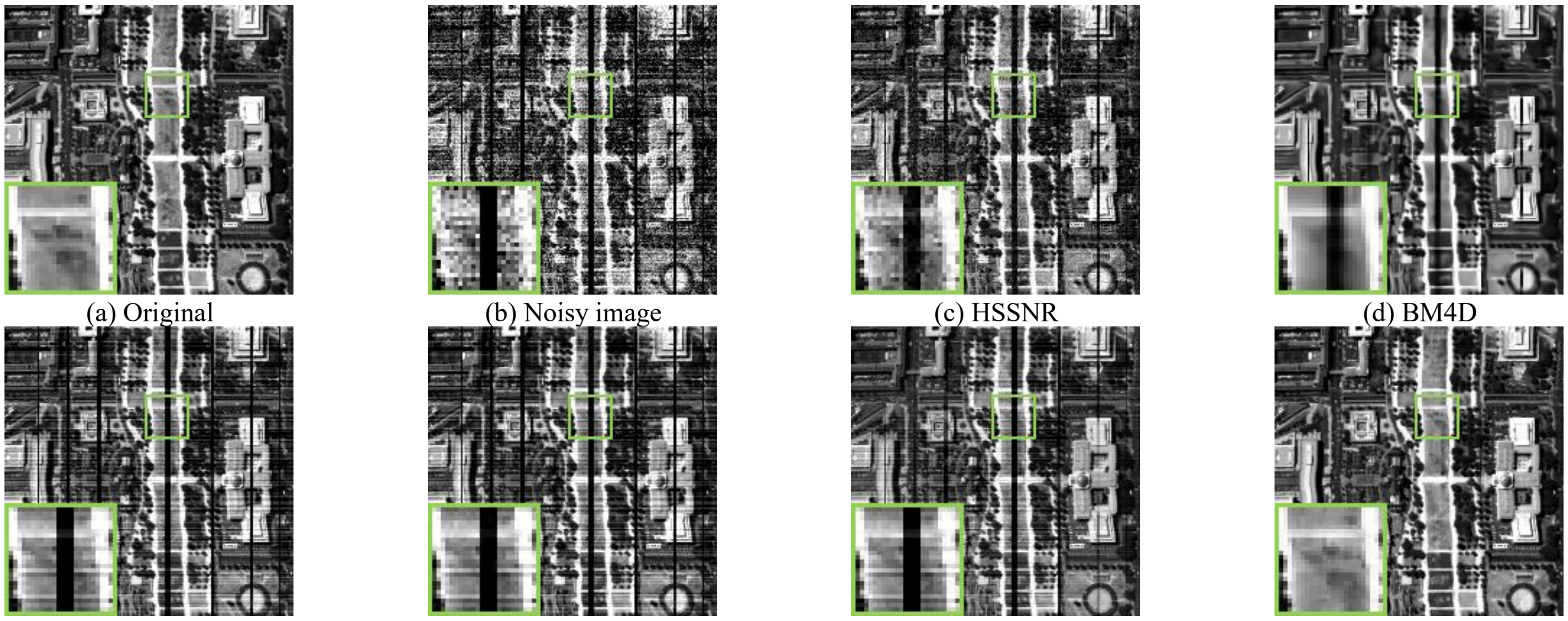

(a) Original (b) Noisy image (c) HSSNR (d) BM4D

(e) LRMR (f) LRTV (g) NonLRMA (h) SSGN

Fig. 8. *Case 5.1*. (a) Original image with band 54. (b) Noisy image. (c) HSSNR. (d) BM4D. (e) LRMR. (f) LRTV. (g) NonLRMA. (h) SSGN.

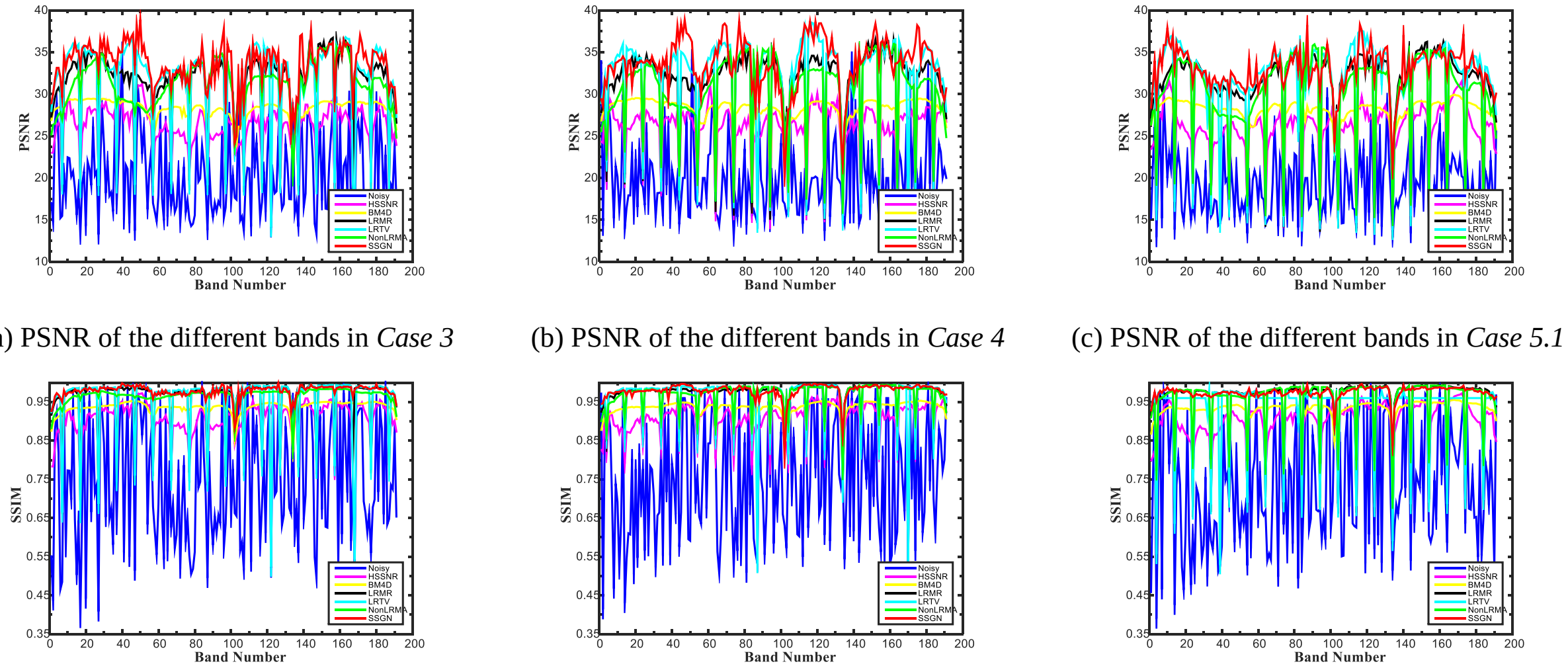

(a) PSNR of the different bands in *Case 3* (b) PSNR of the different bands in *Case 4* (c) PSNR of the different bands in *Case 5.1*

(d) SSIM of the different bands in *Case 3* (e) SSIM of the different bands in *Case 4* (f) SSIM of the different bands in *Case 5.1*

Fig. 9. PSNR and SSIM values of the different denoising methods in each band of the simulated experiments with *Case 3*, *Case 4*, and *Case 5.1*.

***Parameter Sensitivity Analysis:*** The penalty parameter $\alpha$ for the trade-off between the spatial and spectral items in Eq. (7) is critical in the HSI denoising procedure. To explore the influence of $\alpha$ for SSGN, Fig. 10 reveals the quantitative evaluation results (MPSNR and MSA) with different penalty values in the simulated experiments (***Case 1-3***). In ***Case 1*** with only Gaussian noise, the spatial loss function without the spectral item outperforms slightly than the spatial-spectral loss function in the MPSNR results, as shown in Fig. 10(a). The reason may be that the random noise can be effectively described and restrained through just mean square error (MSE) loss. Meanwhile in ***Case 2*** and ***Case 3*** with mixed noise, especially sparse noise, from the perspective of spatial information restoration, the MPSNR results of the proposed SSGN first rise with the increase of $\alpha$ , as shown in Fig. 10 under the stripe or mixed noise scenario, and when the value is equal to 0.001, the results reach the highest MPSNR value. The results then gradually decrease with the increase of $\alpha$ . From the other perspective of spectral information restoration, the MSA results of the proposed SSGN first decrease with the increase of $\alpha$ , as shown in Fig. 10(d), (e), and (f), and when the value is equal to 0.001, the results reach the lowest MSA value. The spectral distortion then gradually rises with the increase of $\alpha$ . Essentially, the spectral preservation strategy is crucial for HSI denoising [45-46], and can simultaneously maintain the spatial structure information and restrain the spectral distortion, especially for sparse noise.

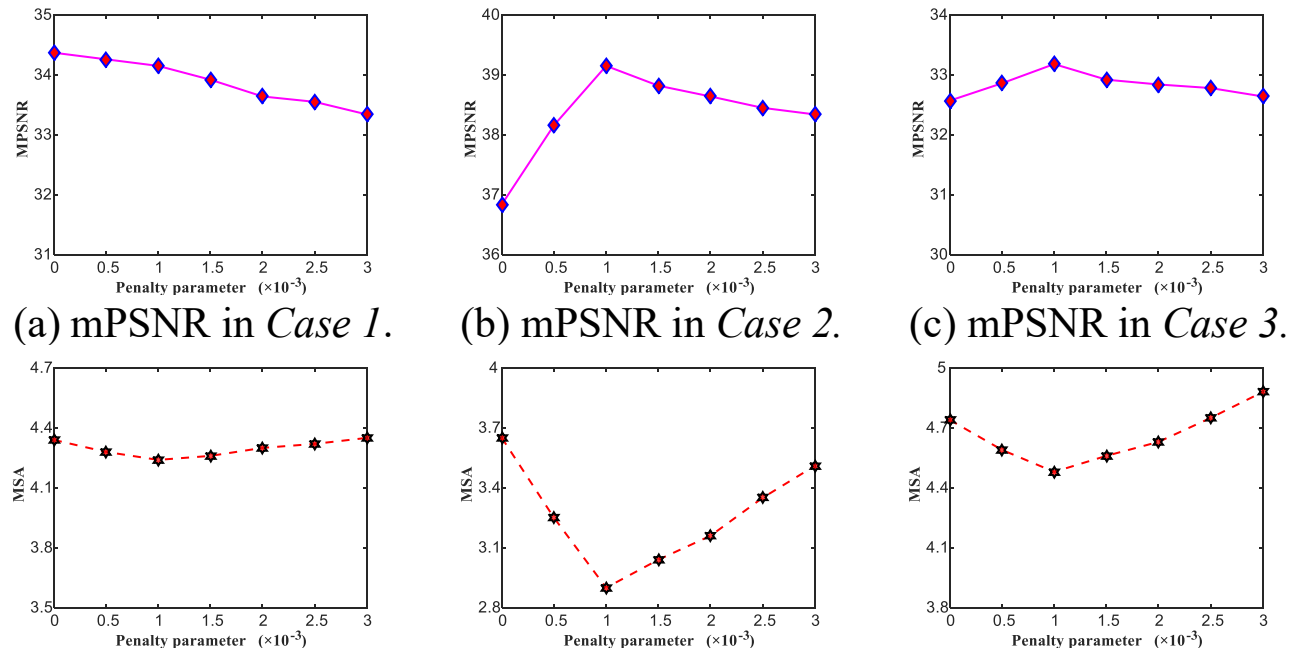

(a) mPSNR in *Case 1*. (b) mPSNR in *Case 2*. (c) mPSNR in *Case 3*.

(d) MSA in *Case 1*. (e) MSA in *Case 2*. (f) MSA in *Case 3*.

Fig. 10. Quantitative evaluation results under different values of $\alpha$ .

## B. Real-Data Experiments

To further test the effectiveness of the proposed SSGN method for HSI mixed noise removal, three real-world HSI data sets, as shown in Figs. 11, 13, and 16, were employed in the real-data experiments. The original and restored mean normalized DN curves, in per-row or column form, through the different methods are given in Figs. 12, 15 and 17, respectively.

**1) HYDICE Urban Data Set:** The noise types are mainly dense noise, stripe noise, and mixed noise of these two types. Fig. 11 displays the denoising results in band number 104 for the five comparing algorithms and the proposed SSGN model, respectively. For a more elaborate comparison, the original and restored mean normalized DN curves, in per-row form, through the different algorithms are shown in Fig. 12.

In Fig. 11, it can be clearly observed that HSSNR can reduce some of the noise, but the mixed noise still remains in the restored results. BM4D does well in suppressing mixed noise, but it also introduces over-smoothing in some regions and loses many high-frequency details due to the NLM strategy. LRMR, LRTV, and NonLRMA all perform well for mixed noise reduction, but they cannot preserve the spectral information well, as shown in Fig. 12. In contrast, SSGN performs the best in Fig. 11(g) and Fig. 12(g), effectively removing the mixed noise, while simultaneously preserving the local details, without introducing obvious over-smoothing or spectral distortion.

**2) AVIRIS Indian Pines Data Set:** The first few bands and several of the middle bands of the Indian Pines HSI are seriously corrupted by Gaussian noise and impulse noise. Figs. 13 and 14 show the denoising results of the different methods, which represent band number 2 and bands (145, 24, 2) of the Indian Pines image, respectively. In Fig. 13, it can be clearly noticed that Gaussian noise and impulse noise still remain in the reconstructed results through HSSNR. BM4D does well in reducing the dense noise, but it appears to be virtually powerless against heavy impulse noise. LRMR and NonLRMA perform well in reducing mixed noise. However, their restored results still exhibit obvious residual noise and stripes. The LRTV method also shows the ability of noise suppression, but some detailed information is simultaneously smoothed and destroyed in Fig. 14(e). SSGN exhibits a best performance for not only effectively removing the dense noise and impulse noise, but also simultaneously preserving the high frequency details and structural information of the Indian Pines image both in Figs. 13-15.

**3) Hyperion EO-1 Data Set:** The first and last few bands of the EO-1 are seriously corrupted by Gaussian noise, stripe noise, and dead lines. Fig. 16, including partial enlarged details, shows the denoising results in band number 2 for five contrast algorithms, and the proposed SSGN, respectively. For a clearer comparison among these methods, the original and restored mean DN curves in per-column through different algorithms are displayed in Fig. 17, respectively.

In Fig. 16, it can be clearly observed that HSSNR can reduce some of the noise, but the mixed noise still remains in the restored results. BM4D does well in suppressing dense noise, but it also introduces over-smoothing in most regions and loses much of the detailed information, by reason of the NLM strategy. LRMR, LRTV, and NonLRMA perform well for mixed noise reduction, but they cannot integrally recover the dead lines, as shown in Fig. 17 and the magnified results in Fig. 16. Outperforming all of the comparison methods, SSGN effectively reduces the mixed noise, while simultaneously preserves the local details and structural information, without bringing obvious over-smoothing effects or spectral distortions.

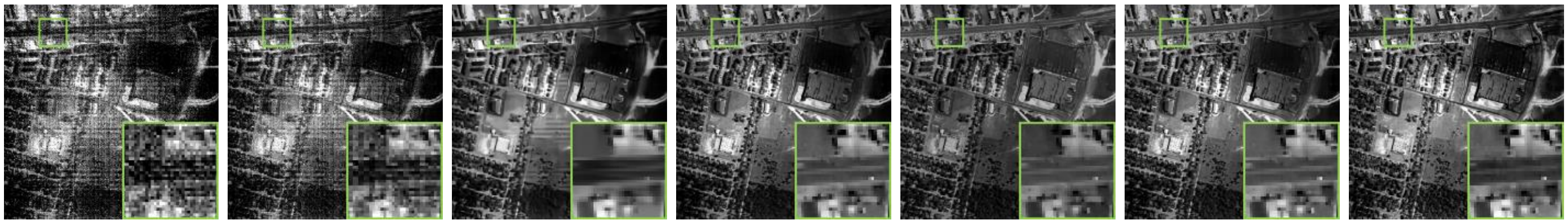

(a) Noisy image (b) HSSNR (c) BM4D (d) LRMR (e) LRTV (f) NonLRMA (g) SSGN

Fig. 11. Denoising results for the HYDICE Urban image. (a) Noisy image band 104. (b) HSSNR. (c) BM4D. (d) LRMR. (e) LRTV. (f) NonLRMA. (g) SSGN.

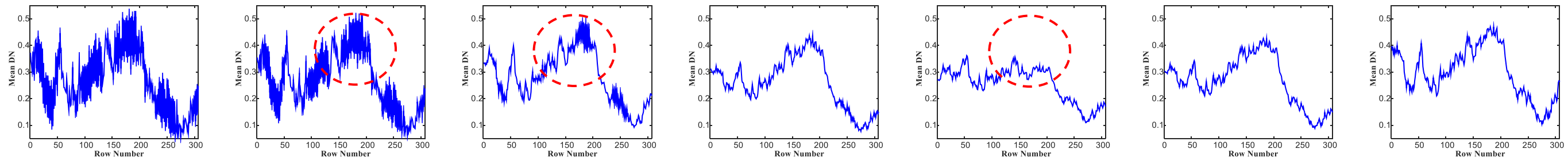

(a) Noisy image (b) HSSNR (c) BM4D (d) LRMR (e) LRTV (f) NonLRMA (g) SSGN

Fig. 12. Horizonal mean DN profiles of band 104 in Urban data set. (a) Noisy image. (b) HSSNR. (c) BM4D. (d) LRMR. (e) LRTV. (f) NonLRMA. (g) SSGN.

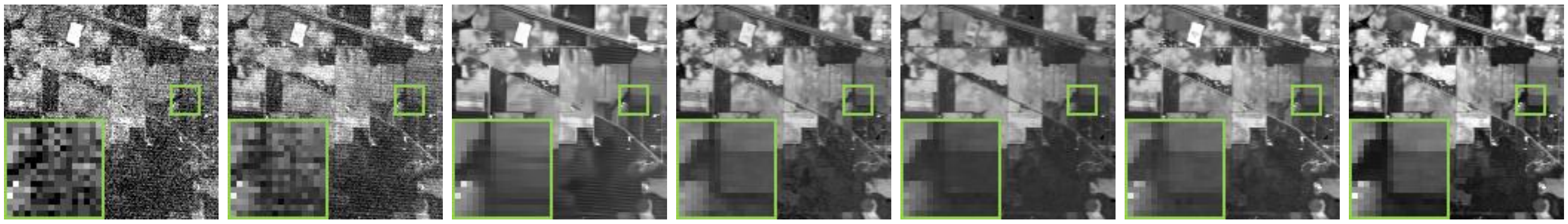

(a) Noisy image (b) HSSNR (c) BM4D (d) LRMR (e) LRTV (f) NonLRMA (g) SSGN

Fig. 13. Denoising Results for the AVIRIS Indian Pines image. (a) Noisy image band 2. (b) HSSNR. (c) BM4D. (d) LRMR. (e) LRTV. (f) NonLRMA. (g) SSGN.

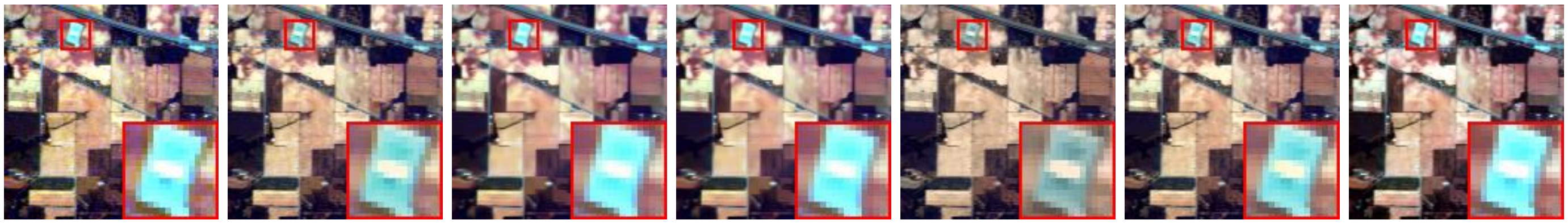

(a) Noisy image (b) HSSNR (c) BM4D (d) LRMR (e) LRTV (f) NonLRMA (g) SSGN

Fig. 14. Denoising Results for Indian Pines image. (a) Noisy image bands (145, 24, 2). (b) HSSNR. (c) BM4D. (d) LRMR. (e) LRTV. (f) NonLRMA. (g) SSGN.

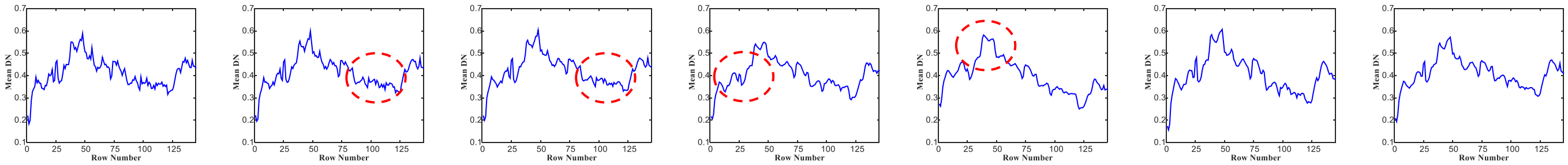

(a) Noisy image (b) HSSNR (c) BM4D (d) LRMR (e) LRTV (f) NonLRMA (g) SSGN

Fig. 15. Horizonal mean DN profiles of band 2 in Indian data set. (a) Noisy image. (b) HSSNR. (c) BM4D. (d) LRMR. (e) LRTV. (f) NonLRMA. (g) SSGN.

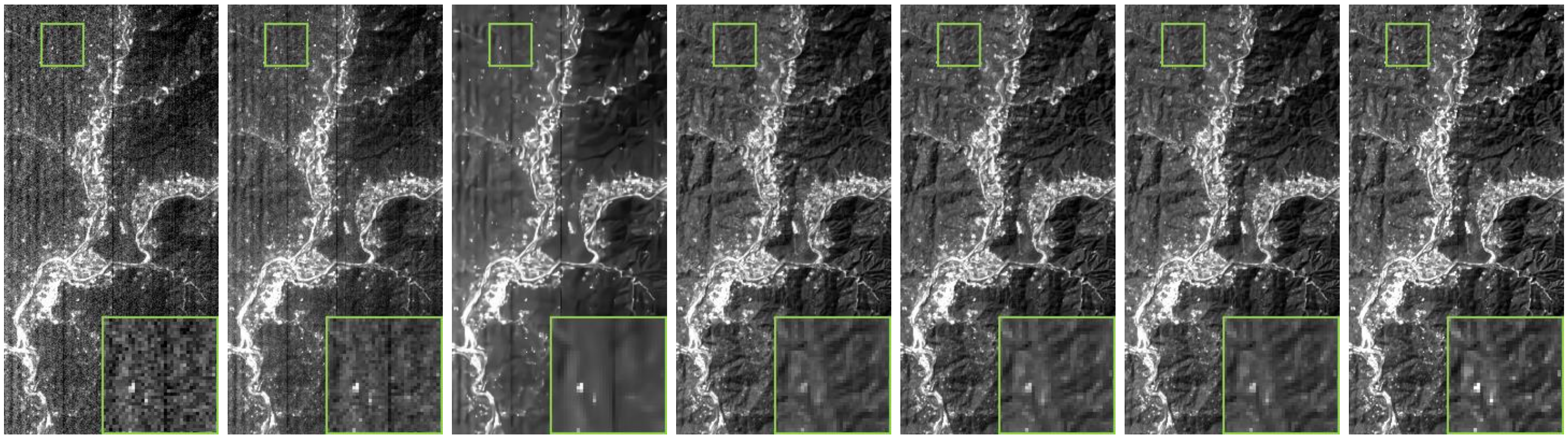

(a) Noisy image (b) HSSNR (c) BM4D (d) LRMR (e) LRTV (f) NonLRMA (g) SSGN

Fig. 16. Denoising Results for the Hyperion EO-1 image. (a) Noisy image band 2. (b) HSSNR. (c) BM4D. (d) LRMR. (e) LRTV. (f) NonLRMA. (g) SSGN.

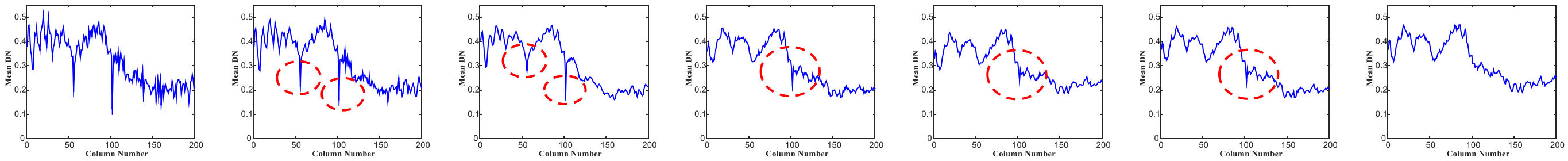

(a) Noisy image (b) HSSNR (c) BM4D (d) LRMR (e) LRTV (f) NonLRMA (g) SSGN

Fig. 17. Vertical mean DN profiles of band 2 in Hyperion EO-1 data set. (a) Noisy image. (b) HSSNR. (c) BM4D. (d) LRMR. (e) LRTV. (f) NonLRMA. (g) SSGN.

### C. *Discussion*

**1) Classification validation:** To further validate the effect of the presented model, the classification results of the Indian Pines image before and after denoising are listed in Fig. 18 with different methods. Support vector machine (SVM) was utilized as the classifier under the same environment for all the restoration results. The overall accuracy (OA) and the kappa coefficient are given as evaluation indexes in Table II. SSGN performs better with the highest OA and kappa indexes of 85.4% and 0.831, respectively. This also verifies the effectiveness of the proposed HSI denoising method.

TABLE II

CLASSIFICATION ACCURACY RESULTS FOR INDIAN PINES.

| | Original | HSSNR | BM4D | LRMR | LRTV | NonLRMA | SSGN |
|---|---|---|---|---|---|---|---|
| OA | 75.9% | 78.7% | 83.9% | 79.4% | 80.5% | 84.7% | **85.4%** |
| Kappa | 0.722 | 0.743 | 0.816 | 0.764 | 0.789 | 0.824 | **0.831** |

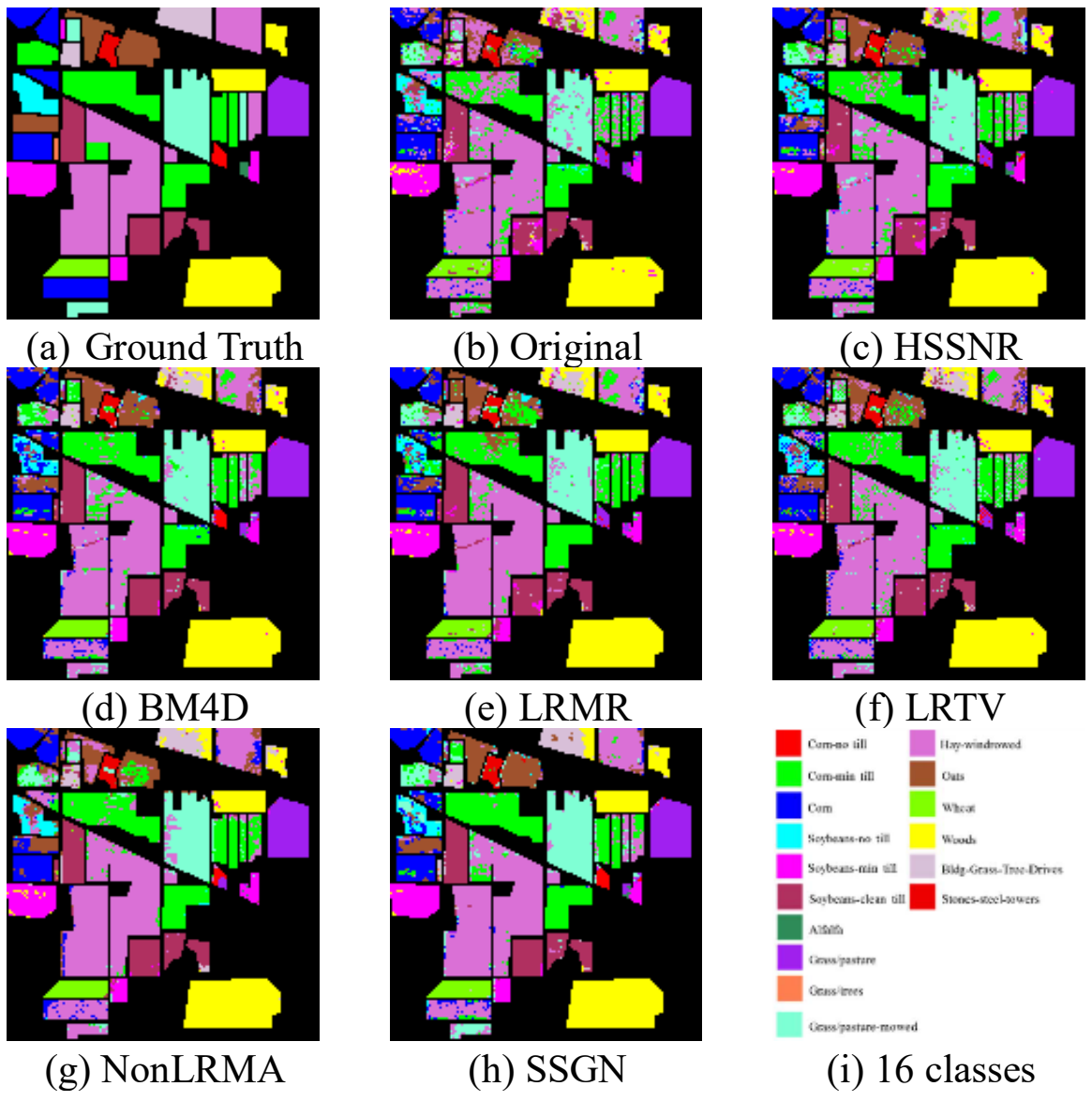

Fig. 18. Classification results for the Indian Pines image.

**2) Run-time comparison:** To compare the work efficiency of the different denoising algorithms, the average running times were recorded for the three real-data experiments under the same operational environment (Software: Windows 7, MATLAB R2014b, CUDA 8.0; Hardware: 16-GB RAM, E5-2609v3 CPU, GTX TITAN X GPU), as listed in Table III. Clearly, the proposed SSGN shows the lowest run-time consumption, due to the high efficiency of this unified learning framework under GPU mode.

TABLE III
RUN-TIME COMPARISON FOR THE REAL-DATA EXPERIMENTS (SECONDS)

| Dataset | HSSNR | BM4D | LRMR | LRTV | NonLRMA | SSGN |
|---|---|---|---|---|---|---|
| Urban | 594.6 | 923.1 | 1147.8 | 385.2 | 1264.3 | **15.6** |
| Indian | 162.7 | 236.3 | 268.7 | 102.4 | 314.8 | **3.9** |
| EO-1 | 486.5 | 812.7 | 936.3 | 317.8 | 972.4 | **11.5** |

**3) Overall evaluation:** Compared with the other existing HSI denoising algorithms in both simulated and actual experiments, SSGN outperforms in most evaluation indices (as listed in Table I), visual assessments (as shown in Figs. 11-17), classification accuracy (as listed in Table II), and time consumption (as listed in Table III), under different mixed noise scenarios. Although the existing HSI denoising methods can obtain favorable results, the inadaptability for hybrid noise removal in different HSIs and the low efficiency issue still restrict the application of HSI denoising. The significant advantage of the proposed model is that SSGN can efficiently deal with multiple types of scenarios, without manually adjusting presupposed parameters through the data-driven mode. Nevertheless, the limitations of the proposed SSGN still exist, such as more complex types of dense-distributed strip noise, and independent band normalization problems.

## V. CONCLUSION

In this work, we present a spatial-spectral gradient network (SSGN) for hybrid noise reduction in HSIs considering the noise type of Gaussian noise, stripe noise, impulse noise, dead line and their mixture. In future works, we will combine more prior characteristics in HSIs with the data-driven learning framework, such as the low-rank tensor factorization model, to effectively utilize the characteristics of HSIs. Besides, holistic bands normalization strategy should be developed for better utilizing redundancy and highly correlated spectral information in HSI denoising task.

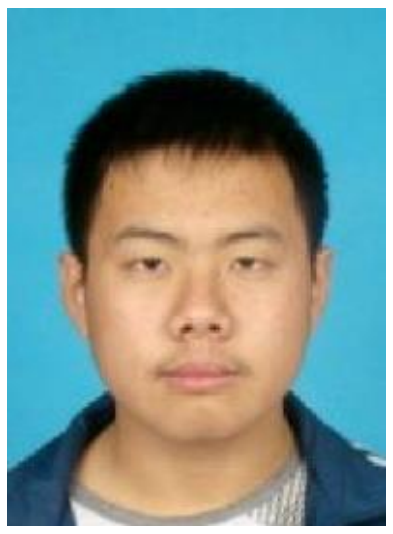

**Qiang Zhang** (S’17) received the B.S. degree in surveying and mapping engineering and M.S. degree in photogrammetry and remote sensing from Wuhan University, Wuhan, China, in 2017 and 2019, respectively. He is currently pursuing the Ph.D. degree in State Key Laboratory of Information Engineering in Surveying, Mapping, and Remote Sensing, Wuhan University, Wuhan, China.

His research interests include image quality improvement, data fusion, machine/deep learning and computer vision.

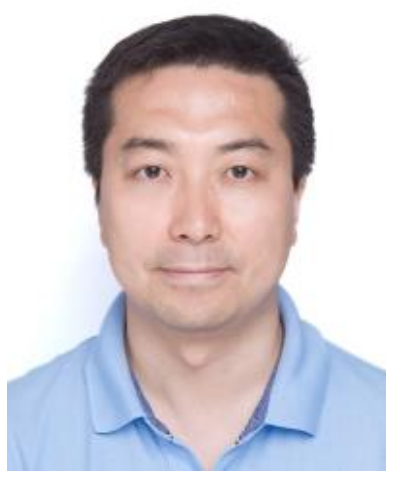

**Qiangqiang Yuan** (M’13) received the B.S. degree in surveying and mapping engineering and the Ph.D. degree in photogrammetry and remote sensing from Wuhan University, Wuhan, China, in 2006 and 2012, respectively.

In 2012, he joined the School of Geodesy and Geomatics, Wuhan University, where he is currently an Associate Professor. He published more than 50 research papers, including more than 30 peer-reviewed articles in international journals such as the IEEE TRANSACTIONS IMAGE PROCESSING and the IEEE TRANSACTIONS ON GEOSCIENCE AND REMOTE SENSING. His current research interests include image reconstruction, remote sensing image processing and application, and data fusion.

Dr. Yuan was the recipient of the Top-Ten Academic Star of Wuhan University in 2011. In 2014, he received the Hong Kong Scholar Award from the Society of Hong Kong Scholars and the China National Postdoctoral Council. He has frequently served as a Referee for more than 20 international journals for remote sensing and image processing.

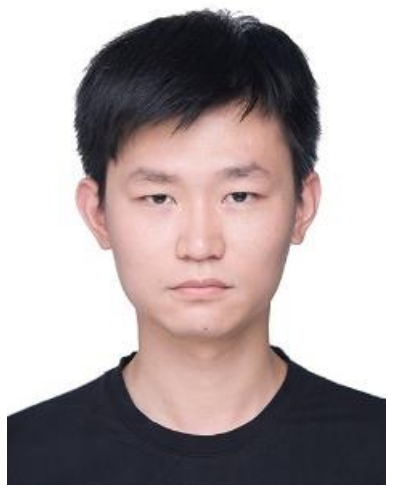

**Jie Li** (M'16) received the B.S. degree in sciences and techniques of remote sensing and the Ph.D. degree in photogrammetry and remote sensing from Wuhan University, Wuhan, China, in 2011 and 2016.

He is currently a Lecturer with the School of Geodesy and Geomatics, Wuhan University. His research interests include image quality improvement, super-resolution, data fusion, remote sensing image processing, sparse representation and deep learning.

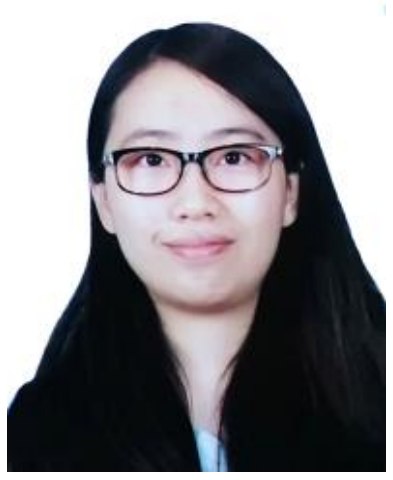

**Xinxin Liu** received the B.S. degree in geographic information system and the Ph.D. degree in Cartography and Geographic Information System from Wuhan University, Wuhan, China, in 2013 and 2018, respectively.

In July 2018, she joined the College of Electrical and Information Engineering, Hunan University, where she is currently an assistant professor. Her research interests include image quality improvement, remote sensing image processing, and remote sensing mapping and application.

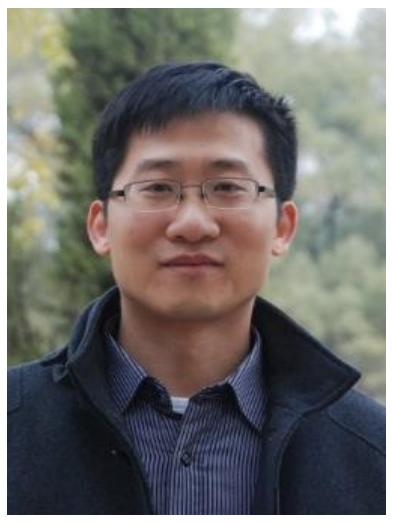

**Huanfeng Shen** (M'10–SM'13) received the B.S. degree in surveying and mapping engineering and the Ph.D. degree in photogrammetry and remote sensing from Wuhan University, Wuhan, China, in 2002 and 2007, respectively.

In 2007, he joined the School of Resource and Environmental Sciences, Wuhan University, where he is currently a Luojia Distinguished Professor. He has been supported by several talent programs, such as the Youth Talent Support Program of China in 2015, the China National Science Fund for Excellent Young Scholars in 2014, and the New Century Excellent Talents by the Ministry of Education of China in 2011. He has authored over 100 research papers. His research interests include image quality improvement, remote sensing mapping and application, data fusion and assimilation, and regional and global environmental changes. Dr. Shen is currently a member of the Editorial Board of the *Journal of Applied Remote Sensing*.

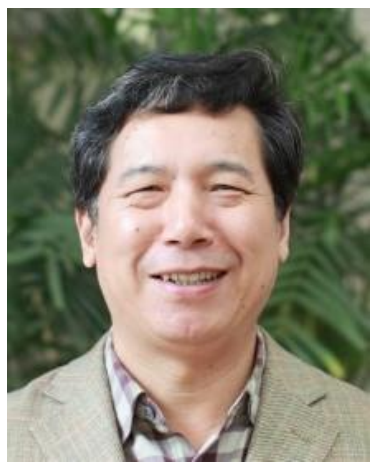

**Liangpei Zhang** (M'06–SM'08–F'19) received the B.S. degree in physics from Hunan Normal University, Changsha, China, in 1982, the M.S. degree in optics from the Xi'an Institute of Optics and Precision Mechanics, Chinese Academy of Sciences, Xi'an, China, in 1988, and the Ph.D. degree in photogrammetry and remote sensing from Wuhan University, Wuhan, China, in 1998.

He was the head of the remote sensing division, State Key Laboratory of Information Engineering in Surveying, Mapping, and Remote Sensing (LIESMARS), Wuhan University. He is also a "Chang-Jiang Scholar" chair professor appointed by the ministry of education of China. He is currently a principal scientist for the China state key basic research project (2011–2016) appointed by the ministry of national science and technology of China to lead the remote sensing program in China. He has more than 600 research papers and seven books. He is the holder of 15 patents. His research interests include hyperspectral remote sensing, high-resolution remote sensing, image processing, and artificial intelligence.

Dr. Zhang is the founding chair of IEEE Geoscience and Remote Sensing Society (GRSS) Wuhan Chapter. He received the best reviewer awards from IEEE GRSS for his service to IEEE Journal of Selected Topics in Earth Observations and Applied Remote Sensing (JSTARS) in 2012 and IEEE Geoscience and Remote Sensing Letters (GRSL) in 2014. He was the General Chair for the 4th IEEE GRSS Workshop on Hyperspectral Image and Signal Processing: Evolution in Remote Sensing (WHISPERS) and the guest editor of JSTARS. His research teams won the top three prizes of the IEEE GRSS 2014 Data Fusion Contest, and his students have been selected as the winners or finalists of the IEEE International Geoscience and Remote Sensing Symposium (IGARSS) student paper contest in recent years.

Dr. Zhang is a Fellow of the Institution of Engineering and Technology (IET), executive member (board of governor) of the China national committee of international geosphere–biosphere programme, executive member of the China society of image and graphics, etc. He was a recipient of the 2010 best paper Boeing award and the 2013 best paper ERDAS award from the American society of photogrammetry and remote sensing (ASPRS). He regularly serves as a Co-chair of the series SPIE conferences on multispectral image processing and pattern recognition, conference on Asia remote sensing, and many other conferences. He edits several conference proceedings, issues, and geoinformatics symposiums. He also serves as an associate editor of the, *International Journal of Image and Graphics, International Journal of Digital Multimedia Broadcasting, Journal of Geo-spatial Information Science, and Journal of Remote Sensing,* and the guest editor of *Journal of applied remote sensing and Journal of sensors.* He is currently serving as an associate editor of the IEEE TRANSACTIONS ON GEOSCIENCE AND REMOTE SENSING.